\documentclass{article}
\usepackage[final]{neurips_2026}

\usepackage[utf8]{inputenc} 
\usepackage[T1]{fontenc}    
\usepackage{hyperref}       
\usepackage{url}            
\usepackage{booktabs}       
\usepackage{graphicx}
\usepackage{multirow}
\usepackage{array}
\usepackage{tabularx}
\usepackage{amsfonts}       
\usepackage{nicefrac}       
\usepackage{microtype}      
\usepackage{xcolor}         
\usepackage{amsmath}
\usepackage{longtable}
\usepackage{placeins}

\title{NoTVLA: Semantics-Preserving Robot Adaptation via Narrative Action Interfaces}

\author{%
\textbf{Zheng Huang$^{*}$ \quad
Mingyu Liu$^{*}$ \quad
Xiaoyi Lin \quad
Muzhi Zhu \quad
Canyu Zhao \quad
Zongze Du}\\
\textbf{Ye Lin \quad
Xiaoman Li \quad
Yiduo Jia \quad
Hao Zhong \quad
Hao Chen$^{\dagger}$ \quad
Chunhua Shen$^{\dagger}$}\\
Zhejiang University\\
\normalfont\footnotesize
$^{*}$Equal contribution.\quad $^{\dagger}$Corresponding author.
}

\begin{document}

\maketitle
\begin{abstract}
Robot fine-tuning can improve manipulation success while degrading the semantic structure inherited from large vision-language pretraining. This trade-off limits the reuse of vision-language-action (VLA) policies in settings that require compositional instructions, camera changes, and coordination with higher-level agents. We present \textbf{NoTVLA}, a semantics-preserving robot adaptation framework that replaces dense low-level action supervision with a sparse narrative action interface. NoTVLA converts demonstrations into sparse, semantically meaningful waypoints, grounds each decision with a task-relevant visual anchor and depth value, and reconstructs executable motion through a deterministic detokenizer. The resulting interface keeps the autoregressive vision-language backbone close to its pretrained prediction format while delegating high-frequency control to a transparent motion-rendering stage. We evaluate this design through matched and backbone-family comparisons, semantic retention probes, semantic out-of-distribution manipulation tasks, camera and depth perturbations, and deployment-oriented efficiency analysis. The results suggest that robot adaptation should be judged not only by task success, but also by how much task-relevant semantic competence is preserved during fine-tuning.
\end{abstract}
\section{Introduction}

Recent progress in vision-language-action (VLA) modeling suggests that large vision-language models (VLMs) provide useful priors for robotic manipulation \citep{driess2023palm, zitkovich2023rt, kim2024openvla, wang2025vq,liu2025bridge}. A persistent tension remains, however: robot fine-tuning can improve task-specific execution while weakening the semantic structure that made the pretrained backbone useful. In practice, a model that initially supports compositional language and open-ended visual concepts may become substantially narrower after being optimized to imitate dense low-level trajectories. As VLM-based policies are increasingly expected to interact with general agent systems, this regression becomes a core limitation rather than a secondary side effect. A policy that succeeds only under a fixed camera setup and a narrow instruction distribution is difficult to reuse, compose, or invoke from higher-level planning systems.

The dominant response has been to strengthen the action side: larger policy heads, diffusion or flow-based decoders, richer control histories, and increasingly specialized robot-side modules \citep{team2024octo, black2024pi_0, chi2025diffusion}. These designs often improve short-horizon control, but they may also widen the gap between semantic reasoning and action generation. Even token-based VLAs are commonly trained to predict dense or near-dense low-level outputs whose structure differs from the narrative and compositional format of language. This mismatch changes the representational pressure placed on the backbone: dense robot supervision can pull the model toward embodiment-specific motor statistics. Recent work treats this preservation problem as a first-class question: VLM2VLA rewrites low-level actions as language and relies on LoRA for adaptation, while MAPS constrains different VLM modules to remain close to their pretrained states through a module-wise proximity schedule \citep{hancock2025actions, huang2025maps}. These methods motivate evaluating robot adaptation by semantic retention as well as execution success.

We take a complementary approach. Rather than attaching a stronger action head to a VLM, we ask whether robot fine-tuning can use a sparse narrative action interface that is closer to language and therefore potentially less disruptive to the pretrained semantic space. We call this framework \textbf{NoTVLA}: not a rejection of VLA modeling, but a test of whether robot adaptation must be centered on dense low-level action prediction. The goal is a semantics-preserving adaptation recipe that keeps the backbone in a narrative decision loop and delegates motion rendering to a lightweight deterministic stage.

\begin{figure}[!htbp]
    \centering
    \IfFileExists{Images/SparseVLA.png}{%
        \includegraphics[width=0.88\textwidth]{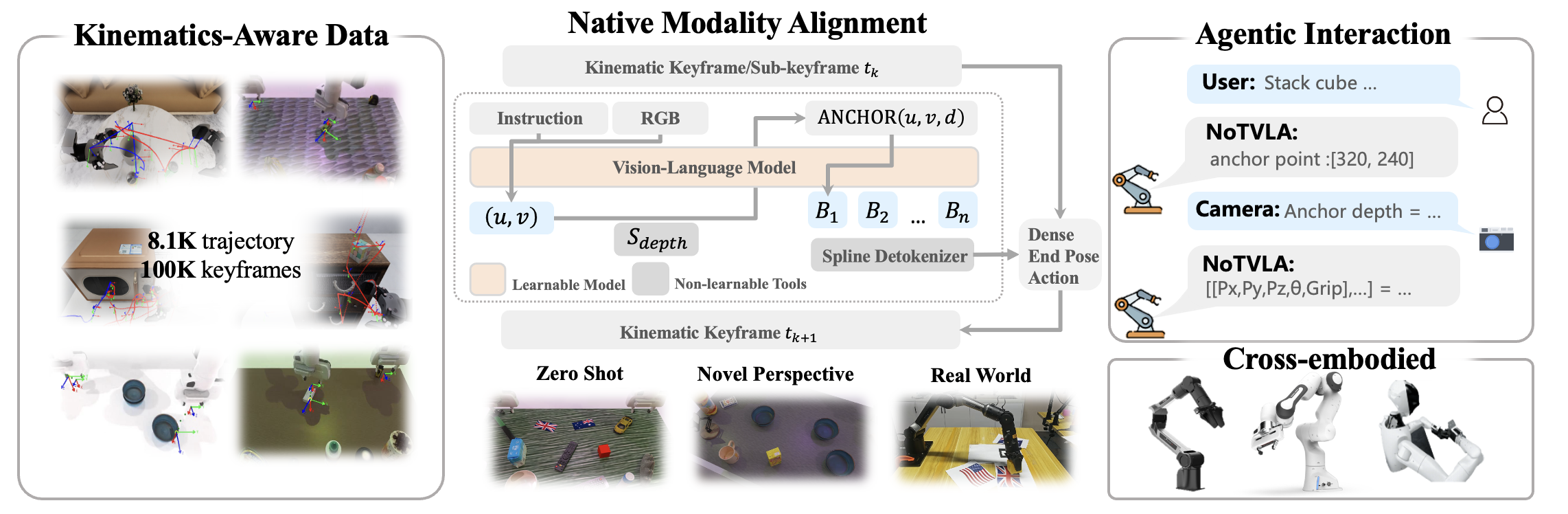}%
    }{%
        \fbox{\parbox{0.88\textwidth}{\centering Placeholder for the NoTVLA system overview.}}%
    }
    \caption{Conceptual overview of NoTVLA. Instead of directly supervising dense control trajectories, the model predicts a task-relevant anchor and a sparse narrative action sequence. A deterministic detokenizer converts these semantically meaningful waypoints into executable robot motion. This shifts adaptation pressure away from embodiment-specific motor statistics and toward the sparse narrative action interface.}
    \label{fig:main}
\end{figure}

NoTVLA is built around three design choices. First, robot demonstrations are compressed into sparse waypoints aligned with semantically meaningful events, such as contact formation, object lifting, and placement completion. Second, the model predicts these waypoints relative to a task-relevant 2D anchor augmented with depth, which preserves geometric grounding without requiring the backbone to regress dense metric trajectories at every step. Third, a deterministic detokenizer reconstructs smooth high-frequency motion from the sparse outputs, separating semantic decision-making from low-level trajectory rendering. In this respect, NoTVLA is closer to recent structured-interface VLAs such as CoT-VLA, HAMSTER, ACoT-VLA, SpatialVLA, and BridgeVLA than to purely dense action predictors. Its emphasis is different, however: the intermediate representation is introduced primarily as a \emph{semantics-preserving adaptation target}, rather than only as a planner, spatial code, or action prior \citep{zhao2025cot, li2025hamster, zhong2026acot, qu2025spatialvla, li2025bridgevla}.

We evaluate NoTVLA as a \emph{semantics-preserving robot adaptation interface}. The claim is intentionally narrower than solving modality mismatch in general. The study focuses on three questions:
\begin{enumerate}
    \item \textbf{Semantic preservation.} Does replacing dense robot supervision with a narrative action interface reduce catastrophic forgetting in the underlying VLM?
    \item \textbf{Generalization.} Is semantic preservation consistently accompanied by stronger performance on semantic inversion, unseen attribute composition, and camera or depth shifts?
    \item \textbf{Practicality.} What are the efficiency, robustness, and deployment trade-offs of such an interface compared with stronger dense-action or specialized-policy alternatives?
\end{enumerate}

We address these questions with matched-protocol comparisons centered on Qwen-PI and Qwen-OFT style baselines where available, backbone-family comparisons, semantic probing, zero-shot composition benchmarks, and deployment-oriented robustness analyses. Public benchmark numbers are used for context, while the semantic retention and representation ablations provide the primary evidence for the mechanism. Additional protocol details, including comparison scope and backbone differences, are provided in the appendix.

Our main contributions are:
\begin{itemize}
    \item We propose \textbf{NoTVLA}, a semantics-preserving robot adaptation framework that replaces dense low-level supervision with a sparse narrative action interface composed of semantic waypoints, anchor-conditioned depth grounding, and deterministic trajectory detokenization.
    \item We identify \emph{semantic preservation during robot adaptation} as a first-class evaluation target for modern VLAs, and provide evidence that the proposed interface retains task-relevant semantic competence more effectively than dense or flow-style robot fine-tuning under matched and backbone-family comparisons.
    \item We provide evidence that the sparse narrative action interface is associated with benefits beyond probe accuracy, including stronger semantic OOD manipulation, improved camera/depth robustness, and a clean separation between low-rate narrative replanning and high-rate continuous execution.
\end{itemize}
\section{Related Works}
\label{sec:related_works}

\subsection{VLA Fine-Tuning and Semantic Retention}
Large VLMs have become the foundation of modern embodied models, from PaLM-E and RT-style systems to open VLAs such as OpenVLA \citep{driess2023palm, zitkovich2023rt, kim2024openvla}. A central challenge in this line of work is that robot adaptation does not simply add a new output head; it also changes the internal representation of the pretrained model. OpenVLA established an open reference point for action-token training and highlighted the fragility of semantic generalization under robot-only fine-tuning \citep{kim2024openvla}. Recent work makes semantic retention an explicit objective. VLM2VLA / Actions-as-Language argues that catastrophic forgetting is rooted in a distribution mismatch between internet-scale VLM pretraining and low-level robot control, and addresses it by rewriting robot actions as natural-language strings so that lightweight LoRA adaptation can retain more VLM competence \citep{hancock2025actions}. MAPS keeps the conventional action supervision format but regularizes transfer at the optimization level, using module-wise proximity scheduling so that visual modules remain close to pretrained priors while action-oriented layers adapt more freely \citep{huang2025maps}. NoTVLA is aligned with this trend but changes the intervention point: rather than verbalizing dense low-level controls or constraining adaptation only through optimization, it changes the supervision target into a sparse narrative action interface with explicit waypoints, anchor-conditioned depth grounding, and deterministic motion rendering.

\subsection{Action Interfaces Beyond Dense Control}
Recent work explores multiple interfaces between semantic reasoning and low-level motion. Some methods discretize or chunk actions for autoregressive prediction \citep{zitkovich2023rt, kim2024openvla}; others use diffusion, flow matching, or latent policy heads to model continuous trajectories \citep{chi2025diffusion, black2024pi_0}. Modular Qwen-based implementations, including StarVLA-$\alpha$, make this comparison cleaner by instantiating chunked regression and flow-expert decoding under a shared backbone, enabling more controlled comparisons between action interfaces \citep{community2026starvla, ye2026starvla}. A third direction focuses on structured intermediate representations, including traces, keyframes, subgoals, paths, and coarse action priors. Actions-as-Language, HAMSTER, and BridgeVLA are representative recent examples: they respectively rewrite low-level actions as language-format supervision, predict coarse 2D end-effector paths before low-level refinement, and align image-space cues with 3D manipulation targets \citep{hancock2025actions, li2025hamster, li2025bridgevla, sun2025learning,wang2026odyssey}. NoTVLA is closest to this broader structured-interface family, but differs in emphasis: the intermediate interface is treated primarily as a \emph{semantics-preserving adaptation target}, not only as a more expressive action representation, path predictor, or spatial grounding module.

\subsection{Trace, Planning, and Narrative Structure}
The abstraction of robot behavior into higher-level traces or plans appears in both classical robotics and recent multimodal foundation models. Hierarchical control decomposes planning and execution across time scales, while embodied foundation models use trace-like signals, foresight, or latent plans to improve generalization. CoT-VLA introduces visual chain-of-thought reasoning by predicting future subgoal images before producing a short action sequence, making the intermediate representation explicitly visual \citep{zhao2025cot}. HAMSTER adopts a hierarchical split in which a high-level VLM proposes a coarse 2D end-effector path and a low-level 3D-aware controller refines it with richer geometry and proprioception \citep{li2025hamster}. This makes HAMSTER a close point of comparison for hierarchical structure, but its emphasis is open-world manipulation through a learned high/low-level split, whereas NoTVLA studies whether the high-level sequence itself can serve as the robot fine-tuning target. ACoT-VLA reasons directly in action space through coarse action intents, implemented with explicit and implicit action reasoners \citep{zhong2026acot}. TraceVLA, Magma, and related systems further suggest that trajectory-like structures can provide an effective interface between visual perception and downstream control \citep{li2025hamster, yang2025magma, liu2025stamo, yang2025learning, su2026world}. NoTVLA uses the notion of narrative more narrowly: the goal is to present robot supervision to the VLM in a format close to instruction-following and semantic disambiguation while delegating smooth motion rendering to a deterministic stage.

\subsection{Spatial Grounding and Depth-Aware Manipulation}
Robotic manipulation requires mapping 2D observations to 3D action decisions. One strategy is to consume richer 3D inputs directly, such as point clouds, voxelized scenes, or ego-centric spatial encodings \citep{song2025reconvla, qu2025spatialvla, liu2025ssr}. SpatialVLA is a representative example: it strengthens the observation side with Ego3D position encoding and redesigns the action side through adaptive action grids, emphasizing spatial representations and discretization rather than an explicit two-stage planner \citep{qu2025spatialvla}. Another route is to remain image-centric but inject task-relevant geometry through keypoints, anchors, depth queries, or affordance maps. BridgeVLA follows this spirit by explicitly aligning VLM image-space outputs with 3D manipulation targets, making input-output alignment the main mechanism for efficient control \citep{li2025bridgevla}. NoTVLA also remains image-centric, but uses the 2D anchor and depth value as one component of the sparse narrative action interface rather than as the full contribution. This choice is not presented as a universal solution to spatial grounding, but as a practical compromise that lets us study semantic retention, camera robustness, and deployment flexibility under a lightweight action interface.

\subsection{Inference Acceleration and Edge Deployment}
Recent work increasingly treats VLA latency as a systems problem rather than a fixed consequence of using large multimodal backbones. Deployment-focused VLA systems push this trend further: LiteVLA-Edge reports fully on-device multimodal control at 150.5\,ms ($\sim$6.6\,Hz) on Jetson AGX Orin through 4-bit quantization and a GPU-accelerated runtime, while NanoVLA argues for reducing latency through cached language computation, long-short action chunking, and dynamic routing \citep{williams2026litevla, chen2025nanovla}. These systems are not direct method baselines for NoTVLA, but they matter for interpretation: a low narrative replanning rate in a current VLM stack should not be confused with an intrinsic limit of sparse narrative action interfaces. Our design is compatible with this deployment trend because high-level narrative inference is already decoupled from continuous execution.
\section{Method}
\label{sec:method}

NoTVLA converts dense robot supervision into a sparse narrative action interface through three stages. Rather than adding a specialized policy head, the framework keeps the autoregressive VLM architecture unchanged and changes the prediction target. The stages are: (1) \textbf{trajectory sparsification}, which compresses demonstrations into semantically meaningful waypoints; (2) \textbf{anchor-conditioned narrative action generation}, which grounds those waypoints in image space and depth; and (3) \textbf{deterministic action detokenization}, which reconstructs executable motion without learning a second low-level controller.

\subsection{Trajectory Sparsification}
\label{sec:keyframe}
The first stage converts dense demonstrations into sparse waypoints that reflect semantic decision points rather than raw control frequency. Instead of supervising every low-level delta action, NoTVLA identifies a compact sequence of events such as contact formation, lift completion, and placement approach. This compression reduces the pressure on the VLM to model embodiment-specific motor rhythms.

Let a raw demonstration be a sequence of end-effector poses $\boldsymbol{T}(t) \in SE(3)$ with translation component $\boldsymbol{p}(t) \in \mathbb{R}^3$ and gripper states $G(t) \in \{0, 1\}$. We identify a set of keyframes $\{t_k\}$ based on kinematic transitions and contact events. A frame at $t_k$ is selected if it satisfies:
\begin{equation}
\label{eq:keyframe_cond}
    \|\ddot{\boldsymbol{p}}(t_k)\|_2 > \alpha \quad \lor \quad \lim_{\epsilon \to 0^+} G(t_k - \epsilon) \neq \lim_{\epsilon \to 0^+} G(t_k + \epsilon),
\end{equation}
where $\alpha$ is a task-dependent translational acceleration threshold. This rule retains decision-critical states while discarding redundant motion. To preserve local geometric context, NoTVLA optionally samples $N$ sub-keyframes between major events, yielding a compact representation $\mathcal{T}_{sparse} = \{\boldsymbol{T}_0, \boldsymbol{T}_1, \dots, \boldsymbol{T}_{M}\}$. The extracted sequence can be variable length; before tokenization, it is resampled to the fixed 5-point waypoint format used by the VLM interface, with endpoint keyframes retained and intermediate points sampled at uniform arc-length intervals.

\subsection{Anchor-Conditioned Narrative Action Generation}
\label{sec:inference}
Directly regressing absolute 3D trajectories from RGB observations forces the model to absorb semantic intent and metric geometry into a single dense output space. NoTVLA instead factorizes grounding into a task-relevant 2D anchor and an associated depth value. This keeps the narrative action representation anchored to the image while avoiding dense 3D prediction at every control step.

\textbf{Anchor Point Prediction (APP).} \enspace
Given an RGB image $\mathbf{I}$ and instruction $\mathbf{l}$, the model first predicts a 2D anchor point $(u_a, v_a)$ representing the projected task locus (e.g., object center or contact point):
\begin{equation}
    (u_a, v_a) = \mathcal{M}_{\text{enc}}(\mathbf{I}, \mathbf{l}), \quad (u_a, v_a) \in [0, W) \times [0, H).
\end{equation}
The depth of this anchor, $d_a$, is retrieved from an external source $\mathcal{S}_{\text{depth}}$ (e.g., a depth sensor or monocular estimator), forming a depth-augmented anchor $a = (u_a, v_a, d_a)$.

\textbf{Narrative Action Sequence.} \enspace
The action sequence is then generated autoregressively conditioned on this anchor:
\begin{equation}
    \mathbf{S} = [\text{CLS}, \mathbf{I}_{\text{emb}}, \mathbf{l}_{\text{emb}}, \text{ANCHOR}(a), B_1, \dots, B_M, \text{EOS}].
\end{equation}
Each action block $B_i$ corresponds to a waypoint in $\mathcal{T}_{sparse}$ and contains discretized depth, image coordinates, gripper state, and rotation tokens:
\begin{equation}
    B_i = (\mathcal{D}_i, \mathcal{U}_i, \mathcal{G}_i, \mathcal{R}_i).
\end{equation}
The model is trained via standard next-token prediction, minimizing the negative log-likelihood over the action tokens:
\begin{equation}
    \mathcal{L}_{\text{VLA}} = - \sum_{t=1}^{|\mathbf{S}|} \log P(s_t | s_{<t}, \mathbf{I}, \mathbf{l}).
\end{equation}
This formulation keeps the backbone unchanged and turns robot adaptation into a next-token prediction problem over a sparse narrative action interface. The model is therefore optimized to predict \emph{what should happen next} in a semantically meaningful sequence, while the continuous realization of that sequence is deferred to the final stage.

\subsection{Deterministic Action Detokenization and Closed-Loop Correction}
\label{sec:detokenization}

The final stage converts sparse narrative outputs into executable motion. NoTVLA uses a deterministic detokenizer rather than a learned low-level policy, allowing the method to separate semantic representation from motion rendering. This design makes the robustness trade-off explicit: improvements must come from the narrative representation and execution-time correction rather than from an additional learned controller.

\textbf{Spline-based Reconstruction.} 
The predicted sparse tokens $\{B_i\}$ are first de-quantized and projected into 3D waypoints $\{\boldsymbol{p}_{w,i}\}$ in the robot base frame using camera intrinsics $\mathbf{K}$ and extrinsics $\boldsymbol{T}_{c}^{w}$. Rotation tokens are de-quantized as Euler angles and converted to unit quaternions before interpolation. NoTVLA constructs a continuous-time trajectory $\hat{\boldsymbol{T}}(t)$ using cubic B-spline \citep{prautzsch2002bezier} interpolation for translation and spherical linear interpolation (SLERP) \citep{shoemake1985animating} for rotation:
\begin{equation}
\begin{split}
    \hat{\boldsymbol{P}}(t) &= \text{Spline}(\{\boldsymbol{p}_{w,i}\}, t), \\
    \hat{\boldsymbol{q}}(t) &= \text{SLERP}(\boldsymbol{q}_i, \boldsymbol{q}_{i+1}, \tau(t)).
\end{split}
\end{equation}
Here $\tau(t)\in[0,1]$ is the normalized time within the active waypoint interval and uses the same monotone knot schedule as the translation spline. This formulation yields smooth interpolation between predicted waypoints and produces an explicit execution trace that can be inspected, constrained, or re-planned.

\textbf{Closed-Loop Trajectory Merging.} 
To support execution-time correction, NoTVLA performs asynchronous secondary inference during motion. After initiating a trajectory $\boldsymbol{T}_0(t)$ at $t_0$, the system triggers a new inference step. When updated waypoints $\mathcal{P}_{new}$ arrive at $t_1$, a merging mechanism blends the new plan with the currently executing trajectory at a future merge point $t_2 = t_1 + \Delta t$.

To prevent kinematic discontinuities, the new plan is aligned with the robot's current state. NoTVLA first identifies the optimal entry point in the new waypoint sequence $\mathcal{P}'=\{(u_i',v_i',d_i')\}_{i=1}^{n'}$ relative to the robot's current end-effector position $\boldsymbol{p}_0(t_2)$, the translation component of $\boldsymbol{T}_0(t_2)$. The optimal index $k^*$ is defined by minimizing the Euclidean distance in the robot base frame:
\begin{equation}
\begin{split}
 k^* = \arg\min_{1\le i\le n'} \big\|\boldsymbol{p}_{w,i}' - \boldsymbol{p}_0(t_2)\big\|_2, \quad
 \boldsymbol{x}_{c,i}' = d_i'\,\mathbf{K}^{-1}[u_i',v_i',1]^\top, \\
 \begin{bmatrix}\boldsymbol{p}_{w,i}' \\ 1\end{bmatrix}
 = \boldsymbol{T}_{c}^{w}
 \begin{bmatrix}\boldsymbol{x}_{c,i}' \\ 1\end{bmatrix}.
\end{split}
\end{equation}
To ensure directional consistency and avoid backtracking, NoTVLA computes a local motion vector $\hat{\boldsymbol{v}}_{\text{path}}$ from the neighboring waypoints of $k^*$ and performs a consistency test:
\begin{equation}
 \hat{\boldsymbol{v}}_{\text{path}} =
 \frac{\boldsymbol{p}_{w,k^+}' - \boldsymbol{p}_{w,k^-}'}
 {\|\boldsymbol{p}_{w,k^+}' - \boldsymbol{p}_{w,k^-}'\|_2}, \quad
 \gamma = (\boldsymbol{p}_{w,k^*}' - \boldsymbol{p}_0(t_2))^\top \hat{\boldsymbol{v}}_{\text{path}},
\end{equation}
where $k^-=\max(1,k^*-1)$ and $k^+=\min(n',k^*+1)$. The waypoint $k^*$ is retained only if $\gamma > 0$. The final executable waypoint sequence discards stale points: $\mathcal{P}_{\text{pending}} = \{(u_j',v_j',d_j')\mid j \ge k^* + \mathbb{1}[\gamma \le 0]\}$. A transition spline then connects $(\boldsymbol{T}_0(t_2), \dot{\boldsymbol{T}}_0(t_2))$ to the start of $\mathcal{P}_{\text{pending}}$, ensuring a smooth handoff. This closed-loop merging procedure enables replanning without requiring the VLM to operate at the actuator control rate.
\section{Experiments}
\label{sec:experiments}

We evaluate NoTVLA along four axes: benchmark performance, semantic retention during robot fine-tuning, robustness of the narrative grounding interface, and real-robot deployment cost. Matched-protocol and backbone-family comparisons provide the main mechanism evidence; public-reference benchmarks and task-level diagnostics appear in Appendix~\ref{app:additional_experiments}.

The comparisons are separated by claim strength. Matched-protocol and backbone-family comparisons support claims about the action interface, while public benchmark numbers provide external context rather than fully controlled head-to-head evidence. The central claim is not that NoTVLA outperforms every VLA implementation, but that a sparse narrative action interface better preserves task-grounded semantics and is associated with stronger semantic OOD manipulation under the tested settings. We report trial counts in the appendix; confidence intervals and multi-seed variance are not yet included, so small differences should be interpreted cautiously.

\subsection{Benchmark Performance}
\label{subsec:main_results}

Table~\ref{tab:main_results} summarizes the main simulator evidence. RoboTwin is the primary matched-protocol comparison \citep{chen2025robotwin}, while SimplerEnv provides public-reference context under the WidowX Visual Matching protocol \citep{li24simpler}. Qwen-PI denotes the Qwen3-VL-4B flow-style policy interface \citep{bai2025qwen3vl}, and Qwen-OFT denotes the Qwen3-VL-4B chunked continuous-action interface. NoTVLA uses Qwen2.5-VL-7B for RoboTwin and Qwen3-VL-4B for SimplerEnv, so the RoboTwin aggregate is not by itself a pure backbone-controlled comparison. We therefore interpret this table together with the semantic retention, semantic OOD, and representation ablations below, where the evidence is closer to the action-interface question.

\begin{table}[!htbp]
\caption{Main benchmark comparison. RoboTwin reports clean/random settings using Easy/Hard naming; Qwen-OFT has Easy-only results. SimplerEnv uses public WidowX Visual Matching averages for context.}
\label{tab:main_results}
\centering
\footnotesize
\setlength{\tabcolsep}{3pt}
\begin{tabularx}{\textwidth}{l l >{\raggedright\arraybackslash}X c}
\toprule
\textbf{Benchmark / Protocol} & \textbf{Method} & \textbf{Interface / Backbone} & \textbf{Success} \\
\midrule
\multicolumn{4}{l}{\textit{RoboTwin 2.0, official 50-clean-trajectory setting (Easy / Hard)}} \\
RoboTwin & $\pi_0$ & VLA + flow expert \citep{black2024pi_0} & 46.42\% / 16.34\% \\
RoboTwin & Qwen-OFT & Qwen3-VL-4B + chunked continuous actions \citep{community2026starvla, ye2026starvla} & 50.38\% (Easy only) \\
RoboTwin & \textbf{NoTVLA (ours)} & Qwen2.5-VL-7B + narrative sparse action & \textbf{63.28\%} / \textbf{28.40\%} \\
\midrule
\multicolumn{4}{l}{\textit{SimplerEnv, WidowX Visual Matching}} \\
SimplerEnv & RT-1-X & large-scale RT-style robot transformer \citep{brohan2022rt} & 1.1\% \\
SimplerEnv & OpenVLA & open token VLA \citep{kim2024openvla} & 4.2\% \\
SimplerEnv & Octo-Base & generalist robot policy \citep{team2024octo} & 16.0\% \\
SimplerEnv & $\pi_0$ & VLA + flow expert \citep{black2024pi_0} & 27.1\% \\
SimplerEnv & $\pi_0$-FAST & tokenized $\pi$ action variant \citep{black2024pi_0, pertsch2025fast} & 48.3\% \\
SimplerEnv & SpatialVLA & PaliGemma2 + spatial tokens \citep{qu2025spatialvla, steiner2024paligemma} & 42.7\% \\
SimplerEnv & GR00T-N1.5 & dual-system foundation policy \citep{bjorck2025gr00t} & 61.9\% \\
SimplerEnv & UniVLA & task-centric latent actions \citep{bu2025univla} & 45.6\% \\
SimplerEnv & SoFar & language-grounded orientation \citep{qi2025sofar} & 58.3\% \\
SimplerEnv & StarVLA-$\pi$ & Qwen3-VL-4B + flow-matching action expert \citep{community2026starvla, ye2026starvla} & 60.9\% \\
SimplerEnv & StarVLA-OFT & Qwen3-VL-4B + chunked continuous actions \citep{community2026starvla, ye2026starvla} & 64.6\% \\
SimplerEnv & StarVLA-GR00T & Qwen3-VL-4B + dual-system action head \citep{community2026starvla, ye2026starvla, bjorck2025gr00t} & 65.3\% \\
SimplerEnv & \textbf{NoTVLA (ours)} & Qwen3-VL-4B + narrative sparse action & \textbf{65.7\%} \\
\bottomrule
\end{tabularx}
\end{table}

\subsection{Semantic Retention and OOD Generalization}
\label{subsec:semantic_preservation}

We first probe whether robot fine-tuning preserves task-grounded language competence. The probe contains 960 examples spanning target identification, attributes, ordering, choice-based placement, and task-stage recognition. This comparison uses the same Qwen3-VL-4B backbone family and evaluation prompt for Qwen-PI and NoTVLA, although the fine-tuning interfaces differ. Qwen-PI rapidly collapses on this probe, while NoTVLA retains 76\% accuracy through the final checkpoint (Table~\ref{tab:semantic_retention}).

\begin{table}[!htbp]
\caption{Task-grounded QA accuracy during robot fine-tuning.}
\label{tab:semantic_retention}
\centering
\small
\begin{tabularx}{\linewidth}{l *{6}{>{\centering\arraybackslash}X}}
\toprule
\textbf{Model} & \textbf{Step 0} & \textbf{Step 100} & \textbf{Step 200} & \textbf{Step 500} & \textbf{Step 1000} & \textbf{Step 4000} \\
\midrule
Qwen-PI & 97\% & 68\% & 23\% & 0\% & 0\% & 0\% \\
\textbf{NoTVLA} & 97\% & 88\% & 79\% & 76\% & 76\% & 76\% \\
\bottomrule
\end{tabularx}
\end{table}

We do not treat the probe alone as causal evidence. It is not designed to reward the NoTVLA output format: the model answers task-grounded semantic questions under a shared prompt, decoding setting, and rule-based scoring protocol. Instead, we use the probe as one part of a matched evidence chain: under comparable robot fine-tuning, the model that retains task-grounded semantics also performs better on semantic OOD manipulation. Table~\ref{tab:semantic_ood} evaluates inverse ordering, unseen color composition, choice-based placement, and novel concept transfer. Dense or flow-style baselines remain competitive on seen tasks, but NoTVLA is substantially better on semantic shifts.

\begin{table}[!htbp]
\caption{Semantic OOD and compositional generalization. OOD Avg. is the unweighted arithmetic mean of the four OOD categories.}
\label{tab:semantic_ood}
\centering
\small
\setlength{\tabcolsep}{4pt}
\begin{tabularx}{\linewidth}{l *{6}{>{\centering\arraybackslash}X}}
\toprule
\textbf{Method} & \textbf{Seen Avg.} & \textbf{Inverse} & \textbf{Novel Color} & \textbf{Choice} & \textbf{Novel Concept} & \textbf{OOD Avg.} \\
\midrule
Qwen-PI-30k & 61.5\% & 0.0\% & 35\% & 23\% & 44\% & 25.5\% \\
Qwen-PI-5k & 50.5\% & 0.0\% & 23\% & 45\% & 59\% & 31.8\% \\
$\pi_0$-30k & 56.5\% & 0.0\% & 30\% & 25\% & 32\% & 21.8\% \\
$\pi_0$-5k & 43.5\% & 0.0\% & 18\% & 56\% & 55\% & 32.3\% \\
$\pi_{0.5}$-30k & 64.0\% & 0.0\% & 48\% & 60\% & 57\% & 41.3\% \\
$\pi_{0.5}$-5k & 52.5\% & 1.0\% & 58\% & 59\% & 62\% & 45.0\% \\
\textbf{NoTVLA}-30k & 57.5\% & 63\% & 62\% & 49\% & 84\% & 64.5\% \\
\textbf{NoTVLA}-5k & 55.0\% & 70\% & 69\% & 62\% & 78\% & 69.8\% \\
\bottomrule
\end{tabularx}
\end{table}

\subsection{Anchor and Depth Grounding Ablation}
\label{subsec:camera_depth}

NoTVLA grounds sparse narrative actions through an image anchor and its depth value. Table~\ref{tab:camera_depth} isolates this design while keeping the policy family fixed.

\begin{table}[!htbp]
\caption{Ablation of anchor-conditioned depth grounding.}
\label{tab:camera_depth}
\centering
\small
\setlength{\tabcolsep}{3.5pt}
\begin{tabularx}{\linewidth}{>{\raggedright\arraybackslash}X c c c}
\toprule
\multirow{2}{*}{\textbf{Task}} & \textbf{Full NoTVLA} & \textbf{Direct UVD} & \textbf{APP only} \\
& \textbf{(APP + depth)} & \textbf{(w/o anchor)} & \textbf{(no depth)} \\
\midrule
\multicolumn{4}{l}{\textit{\textbf{Object Recognition}}} \\
Place a2b left       & \textbf{0.38} & 0.28 & 0.30 \\
Place a2b right      & \textbf{0.34} & 0.24 & 0.32 \\
Place object scale   & \textbf{0.43} & 0.39 & 0.36 \\
\midrule
\multicolumn{4}{l}{\textit{\textbf{Long-Horizon Tasks}}} \\
Blocks ranking size  & \textbf{0.53} & 0.33 & 0.28 \\
Blocks ranking rgb   & \textbf{0.76} & 0.66 & 0.59 \\
Stack block three    & \textbf{0.49} & 0.32 & 0.30 \\
Stack bowl three     & \textbf{0.83} & 0.78 & 0.68 \\
\bottomrule
\end{tabularx}
\end{table}

Full anchor-conditioned grounding improves the seven-task average from 42.9\% for direct UVD and 40.4\% for anchor-point-only prediction to 53.7\%. The gain is largest on long-horizon tasks, where repeated localization makes stable grounding important. Table~\ref{tab:detokenizer_quality_main} shows that the 5-point deterministic detokenizer remains close to the stronger 10-point variant while keeping actions shorter.

\begin{table}[!htbp]
\caption{Compact trajectory-rendering diagnostics for the deterministic detokenizer. Full diagnostics are reported in Appendix~\ref{sec:ablations_appendix}.}
\label{tab:detokenizer_quality_main}
\centering
\scriptsize
\resizebox{\linewidth}{!}{
\begin{tabular}{lccccc}
\toprule
\textbf{Metric} & \textbf{NoTVLA 5-pt} & \textbf{NoTVLA 10-pt} & \textbf{Magma} & \textbf{HAMSTER} & \textbf{LLARVA} \\
\midrule
Cover F1 $\uparrow$ & 0.9546 & \textbf{0.9567} & 0.9282 & 0.9451 & 0.9189 \\
DTW $\downarrow$ & 0.4564 & \textbf{0.4554} & 0.8353 & 0.7945 & 0.4894 \\
Endpoint error $\downarrow$ & 0.0739 & \textbf{0.0729} & 0.1784 & 0.1181 & 0.0771 \\
LCSS similarity $\uparrow$ & 0.9660 & \textbf{0.9671} & 0.9562 & 0.9496 & 0.9447 \\
\bottomrule
\end{tabular}
}
\end{table}

\subsection{Camera and Depth Stress Tests}
\label{subsec:depth_stress}

We also test a precision card-grasping task. The hard split fails if the card contacts the groove boundary, making it more sensitive to depth error than ordinary grasp success.

\begin{table}[!htbp]
\caption{Depth-sensitive grasping. Simulated settings use 100 trials; real paper drawing uses 20 trials. DP3 refers to 3D Diffusion Policy \citep{ze20243d}.}
\label{tab:depth_sensitive_grasp}
\centering
\small
\setlength{\tabcolsep}{6pt}
\begin{tabular}{lccc}
\toprule
\textbf{Method} & \textbf{Sim Card Easy} & \textbf{Sim Card Hard} & \textbf{Real Draw Paper} \\
\midrule
DP3 & 0.67 & 0.18 & 0.20 \\
Qwen-PI & \textbf{0.80} & 0.26 & 0.40 \\
NoTVLA w/ direct UVD & 0.73 & 0.16 & 0.35 \\
\textbf{NoTVLA} & 0.75 & \textbf{0.56} & \textbf{0.45} \\
\bottomrule
\end{tabular}
\end{table}

The hard split exposes the difference between direct depth-token prediction and anchor-conditioned depth use: direct UVD drops to 16\%, while full NoTVLA reaches 56\%. With injected depth-anchor noise, average success changes from 42\% at 0\,cm to 30\% at 1\,cm, 11\% at 10\,cm, and 7\% at 100\,cm; under mild fisheye distortion, anchor-depth grounding averages 27.8\% versus 19.8\% for direct UVD (Appendix~\ref{sec:depth_noise}). These stress tests clarify the failure mode: NoTVLA improves robustness relative to direct UVD prediction, but still needs reliable local geometry and degrades sharply under large depth corruption.

\begin{figure}[!htbp]
    \centering
    \IfFileExists{Images/close_loop_in_grasp.png}{%
        \includegraphics[width=0.92\linewidth]{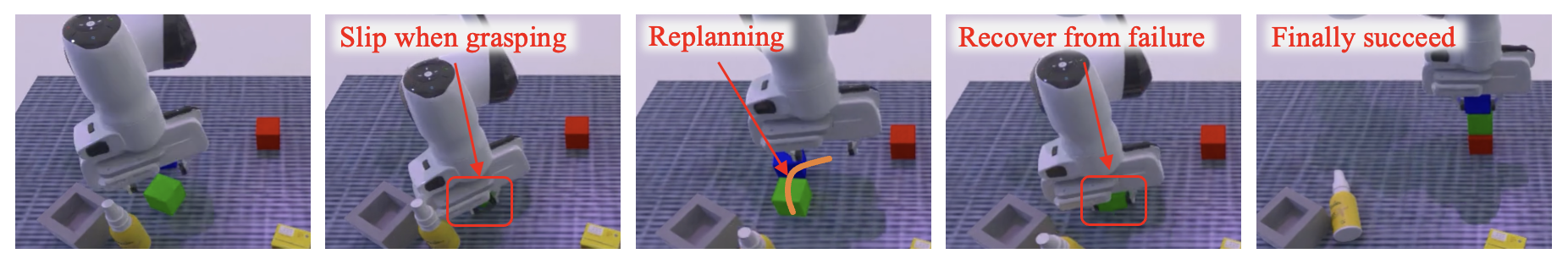}%
    }{%
        \fbox{\parbox{0.85\linewidth}{\centering Placeholder for a closed-loop recovery example.}}%
    }
    \caption{Closed-loop recovery from an initial grasping error through re-planning.}
    \label{fig:regrasp}
\end{figure}

\subsection{Deployment Cost and Real-Robot Results}
\label{subsec:efficiency}

NoTVLA separates low-rate narrative replanning from high-rate execution. In our current implementation, narrative plans are refreshed at 0.25--0.5\,Hz (a 2--4\,s high-level replanning period) with roughly 13\,GB peak inference memory, while deterministic interpolation executes at 80\,Hz. High-level latency therefore affects how often the policy rethinks, not the actuator control frequency. We view this replanning rate as a systems bottleneck rather than an intrinsic limit of the interface.

We deploy the system on five real-robot tasks: pressing a button, placing a cube on a flag, stacking two cubes, inserting a flower, and picking fruit into a bowl. This limited study is a deployment sanity check rather than primary evidence for broad real-world generalization. The flower task fails if the flower falls; other tasks follow Appendix~\ref{app:deployment}. Table~\ref{tab:real_robot_summary} reports averages, while Table~\ref{tab:real_robot_task_success} gives task-level success.

\begin{table}[!htbp]
\caption{Real-robot summary over five manipulation tasks. Time is averaged over successful trials.}
\label{tab:real_robot_summary}
\centering
\small
\setlength{\tabcolsep}{6pt}
\begin{tabularx}{0.72\textwidth}{>{\raggedright\arraybackslash}X c c c}
\toprule
\textbf{Method} & \textbf{Tasks} & \textbf{Success Avg.} & \textbf{Avg. Time} \\
\midrule
DP3 & 5 & 44\% & 45.3s \\
Qwen-PI & 5 & 49\% & 54.7s \\
NoTVLA w/ direct UVD & 5 & 45\% & 37.5s \\
\textbf{NoTVLA} & 5 & \textbf{53\%} & 52.3s \\
\bottomrule
\end{tabularx}
\end{table}

\begin{table}[!htbp]
\caption{Real-robot task-level success. Each task uses task-specific termination criteria.}
\label{tab:real_robot_task_success}
\centering
\scriptsize
\resizebox{\linewidth}{!}{
\begin{tabular}{lccccc}
\toprule
\textbf{Method} & \textbf{Press Button} & \textbf{Place Cube Flag} & \textbf{Stack Cube Two} & \textbf{Insert Flower} & \textbf{Pick Fruit Bowl} \\
\midrule
DP3 & 80\% & 40\% & 30\% & 5\% & \textbf{65\%} \\
Qwen-PI & 85\% & \textbf{70\%} & 25\% & \textbf{10\%} & 55\% \\
NoTVLA w/ direct UVD & 75\% & 55\% & 30\% & 5\% & 60\% \\
\textbf{NoTVLA} & \textbf{90\%} & 60\% & \textbf{50\%} & \textbf{10\%} & 55\% \\
\bottomrule
\end{tabular}
}
\end{table}

Despite slower high-level inference, NoTVLA has comparable successful-trial execution time and the highest average real-robot success. This supports treating narrative replanning rate and actuator control rate as separate deployment quantities.
\section{Conclusion}
\label{sec:conclusion}

This paper presents NoTVLA as a semantics-preserving robot adaptation strategy rather than a generic solution to action-language mismatch. Under the tested protocols, robot fine-tuning through a sparse narrative action interface appears to retain more of the VLM backbone's task-relevant semantic structure and is associated with better generalization beyond dense demonstration distributions.

The evaluation tests this claim through matched and backbone-family comparisons, semantic retention probes, semantic OOD generalization, camera and depth robustness, detokenizer diagnostics, and deployment metrics. These results support evaluating robot adaptation not only by execution success, but also by whether it preserves useful semantic competence.

More broadly, robot adaptation should be evaluated by whether the resulting policy remains compositional, reusable, and compatible with general agentic systems. NoTVLA is one concrete step toward that broader view.

Because NoTVLA decouples high-level narrative inference from low-level execution, faster multimodal runtimes can improve replanning without changing the interface.

\bibliographystyle{plainnat}
\bibliography{ref}


\clearpage
\appendix
\section{Appendix Overview}
\label{app:overview}

This appendix mirrors the main-paper evidence chain. Because the NoTVLA evaluation is centered on RoboTwin and SimplerEnv, we prioritize those two simulators in the appendix tables as well. We also include depth, camera, ablation, and real-robot diagnostics that support the main experimental discussion.

\section{Additional Experimental Results}
\label{app:additional_experiments}

\subsection{Full RoboTwin Results}
\label{sec:full_robotwin}

\begingroup
\scriptsize
\setlength{\tabcolsep}{2pt}
\setlength{\LTleft}{0pt}
\setlength{\LTright}{0pt}
\begin{longtable}{@{}p{0.31\linewidth}*{8}{>{\centering\arraybackslash}p{0.077\linewidth}}@{}}
\caption{Public RoboTwin 2.0 \textbf{Easy} task-level results \citep{chen2025robotwin}. To reduce width, the RoboTwin appendix is split into separate Easy and Hard tables. The Qwen-OFT column reports the Easy-only run available in this setting; its average comes from the full run snapshot. Baseline families include RDT \citep{liu2024rdt}, ACT \citep{zhao2023learning}, DP3 \citep{ze20243d}, and Qwen-style StarVLA-$\alpha$ variants \citep{community2026starvla, ye2026starvla}. Rows with dashes indicate entries unavailable for that method. Public baseline and NoTVLA averages use the 50-task sheet; Qwen-OFT reports the available Easy-only run.} \label{tab:full_robotwin_easy} \\
\toprule
\textbf{Task} & \textbf{RDT} & \textbf{$\pi_0$} & \textbf{$\pi_{0.5}$} & \textbf{ACT} & \textbf{DP} & \textbf{DP3} & \textbf{Qwen-OFT} & \textbf{NoTVLA} \tabularnewline
\midrule
\endfirsthead
\multicolumn{9}{c}{{\bfseries \tablename\ \thetable{} -- continued from previous page}} \\
\toprule
\textbf{Task} & \textbf{RDT} & \textbf{$\pi_0$} & \textbf{$\pi_{0.5}$} & \textbf{ACT} & \textbf{DP} & \textbf{DP3} & \textbf{Qwen-OFT} & \textbf{NoTVLA} \tabularnewline
\midrule
\endhead
\bottomrule
\multicolumn{9}{r}{{Continued on next page}} \\
\endfoot
\bottomrule
\endlastfoot
Adjust Bottle & 81\% & 90\% & 97\% & 97\% & 97\% & 99\% & 96\% & 93\% \\
Beat Block Hammer & 77\% & 43\% & 3\% & 56\% & 42\% & 72\% & 58\% & 93\% \\
Blocks Ranking RGB & 3\% & 19\% & 63\% & 1\% & 0\% & 3\% & 45\% & 76\% \\
Blocks Ranking Size & 0\% & 7\% & 33\% & 0\% & 1\% & 2\% & 27\% & 49\% \\
Click Alarmclock & 61\% & 63\% & 22\% & 32\% & 61\% & 77\% & 91\% & 78\% \\
Click Bell & 80\% & 44\% & 6\% & 58\% & 54\% & 90\% & 94\% & 94\% \\
Dump Bin Bigbin & 64\% & 83\% & 97\% & 68\% & 49\% & 85\% & 68\% & 23\% \\
Grab Roller & 74\% & 96\% & 100\% & 94\% & 98\% & 98\% & 93\% & 97\% \\
Handover Block & 45\% & 45\% & 42\% & 42\% & 10\% & 70\% & 0\% & 55\% \\
Handover Mic & 90\% & 98\% & 37\% & 85\% & 53\% & 100\% & 39\% & 99\% \\
Hanging Mug & 23\% & 11\% & 19\% & 7\% & 8\% & 17\% & 15\% & 15\% \\
Lift Pot & 72\% & 84\% & 0\% & 88\% & 39\% & 97\% & 0\% & 95\% \\
Move Can Pot & 25\% & 58\% & 59\% & 22\% & 39\% & 70\% & 50\% & 76\% \\
Move Pillbottle Pad & 8\% & 21\% & 66\% & 0\% & 1\% & 41\% & 54\% & 12\% \\
Move Playingcard Away & 43\% & 53\% & 87\% & 36\% & 47\% & 68\% & 69\% & 77\% \\
Move Stapler Pad & 2\% & 0\% & 8\% & 0\% & 1\% & 12\% & 12\% & 30\% \\
Open Laptop & 59\% & 85\% & 14\% & 56\% & 49\% & 82\% & 31\% & 62\% \\
Open Microwave & 37\% & 80\% & 18\% & 86\% & 5\% & 61\% & \textemdash & 34\% \\
Pick Diverse Bottles & 2\% & 27\% & 83\% & 7\% & 6\% & 52\% & 30\% & 77\% \\
Pick Dual Bottles & 42\% & 57\% & 86\% & 31\% & 24\% & 60\% & 43\% & 88\% \\
Place A2B Left & 3\% & 31\% & 64\% & 1\% & 2\% & 46\% & 20\% & 38\% \\
Place A2B Right & 1\% & 27\% & 59\% & 0\% & 13\% & 49\% & 22\% & 34\% \\
Place Bread Basket & 10\% & 17\% & 60\% & 6\% & 14\% & 26\% & 52\% & 63\% \\
Place Bread Skillet & 5\% & 23\% & 59\% & 7\% & 11\% & 19\% & 56\% & 67\% \\
Place Burger Fries & 50\% & 80\% & 66\% & 49\% & 72\% & 72\% & 96\% & 69\% \\
Place Can Basket & 19\% & 41\% & 53\% & 1\% & 18\% & 67\% & 63\% & 77\% \\
Place Cans Plasticbox & 6\% & 34\% & 28\% & 16\% & 40\% & 48\% & 81\% & 51\% \\
Place Container Plate & 78\% & 88\% & 90\% & 72\% & 41\% & 86\% & 99\% & 92\% \\
Place Dual Shoes & 4\% & 15\% & 46\% & 9\% & 8\% & 13\% & 28\% & 21\% \\
Place Empty Cup & 56\% & 37\% & 96\% & 61\% & 37\% & 65\% & 72\% & 90\% \\
Place Fan & 12\% & 20\% & 45\% & 1\% & 3\% & 36\% & 28\% & 30\% \\
Place Mouse Pad & 1\% & 7\% & 40\% & 0\% & 0\% & 4\% & 9\% & 23\% \\
Place Object Basket & 33\% & 16\% & 67\% & 15\% & 15\% & 65\% & 40\% & 76\% \\
Place Object Scale & 1\% & 10\% & 73\% & 0\% & 1\% & 15\% & 19\% & 43\% \\
Place Object Stand & 15\% & 36\% & 80\% & 1\% & 22\% & 60\% & 48\% & 75\% \\
Place Phone Stand & 15\% & 35\% & 54\% & 2\% & 13\% & 44\% & 24\% & 28\% \\
Place Shoe & 35\% & 28\% & 77\% & 5\% & 23\% & 58\% & 63\% & 59\% \\
Press Stapler & 41\% & 62\% & 61\% & 31\% & 6\% & 69\% & 60\% & 94\% \\
Put Bottles Dustbin & 21\% & 54\% & 63\% & 27\% & 22\% & 60\% & \textemdash & 55\% \\
Put Object Cabinet & 33\% & 68\% & 34\% & 15\% & 42\% & 72\% & 35\% & 84\% \\
Rotate QRcode & 50\% & 68\% & 80\% & 1\% & 13\% & 74\% & 50\% & 73\% \\
Scan Object & 4\% & 18\% & 40\% & 2\% & 9\% & 31\% & 13\% & 24\% \\
Shake Bottle Horizontally & 84\% & 99\% & 100\% & 63\% & 59\% & 100\% & 98\% & 96\% \\
Shake Bottle & 74\% & 97\% & 100\% & 74\% & 65\% & 98\% & 98\% & 94\% \\
Stack Blocks Three & 2\% & 17\% & 54\% & 0\% & 0\% & 1\% & 41\% & 49\% \\
Stack Blocks Two & 21\% & 42\% & 85\% & 25\% & 7\% & 24\% & 83\% & 88\% \\
Stack Bowls Three & 51\% & 66\% & 70\% & 48\% & 63\% & 57\% & 62\% & 83\% \\
Stack Bowls Two & 76\% & 91\% & 93\% & 82\% & 61\% & 83\% & 90\% & 96\% \\
Stamp Seal & 1\% & 3\% & 48\% & 2\% & 2\% & 18\% & 27\% & 36\% \\
Turn Switch & 35\% & 27\% & 25\% & 5\% & 36\% & 46\% & 26\% & 33\% \\
\midrule
Average & 34.50\% & 46.42\% & 57.00\% & 29.74\% & 28.04\% & 55.24\% & 50.38\% & 63.28\% \\
\end{longtable}
\endgroup
\FloatBarrier

\begingroup
\scriptsize
\setlength{\tabcolsep}{2pt}
\setlength{\LTleft}{0pt}
\setlength{\LTright}{0pt}
\begin{longtable}{@{}p{0.31\linewidth}*{7}{>{\centering\arraybackslash}p{0.088\linewidth}}@{}}
\caption{Public RoboTwin 2.0 \textbf{Hard} task-level results \citep{chen2025robotwin}. To reduce width, the RoboTwin appendix is split into separate Easy and Hard tables. The $\pi_{0.5}$ column is added from the public \texttt{pi0.5} release. Rows with dashes indicate entries unavailable for that method. Public baseline and NoTVLA averages use the 50-task sheet.} \label{tab:full_robotwin_hard} \\
\toprule
\textbf{Task} & \textbf{RDT} & \textbf{$\pi_0$} & \textbf{$\pi_{0.5}$} & \textbf{ACT} & \textbf{DP} & \textbf{DP3} & \textbf{NoTVLA} \\
\midrule
\endfirsthead
\multicolumn{8}{c}{{\bfseries \tablename\ \thetable{} -- continued from previous page}} \\
\toprule
\textbf{Task} & \textbf{RDT} & \textbf{$\pi_0$} & \textbf{$\pi_{0.5}$} & \textbf{ACT} & \textbf{DP} & \textbf{DP3} & \textbf{NoTVLA} \\
\midrule
\endhead
\bottomrule
\multicolumn{8}{r}{{Continued on next page}} \\
\endfoot
\bottomrule
\endlastfoot
Adjust Bottle & 75\% & 56\% & 54\% & 23\% & 0\% & 3\% & 56\% \\
Beat Block Hammer & 37\% & 21\% & 0\% & 3\% & 0\% & 8\% & 18\% \\
Blocks Ranking RGB & 0\% & 5\% & 8\% & 0\% & 0\% & 0\% & 12\% \\
Blocks Ranking Size & 0\% & 1\% & 11\% & 0\% & 0\% & 0\% & 26\% \\
Click Alarmclock & 12\% & 11\% & 62\% & 4\% & 5\% & 14\% & 27\% \\
Click Bell & 9\% & 3\% & 85\% & 3\% & 0\% & 0\% & 33\% \\
Dump Bin Bigbin & 32\% & 24\% & 30\% & 1\% & 0\% & 53\% & 12\% \\
Grab Roller & 43\% & 80\% & 63\% & 25\% & 0\% & 2\% & 36\% \\
Handover Block & 14\% & 8\% & 0\% & 0\% & 0\% & 0\% & 42\% \\
Handover Mic & 31\% & 13\% & 2\% & 0\% & 0\% & 3\% & 27\% \\
Hanging Mug & 16\% & 3\% & 1\% & 0\% & 0\% & 1\% & 11\% \\
Lift Pot & 9\% & 36\% & 0\% & 0\% & 0\% & 0\% & 49\% \\
Move Can Pot & 12\% & 21\% & 0\% & 4\% & 0\% & 6\% & 58\% \\
Move Pillbottle Pad & 0\% & 1\% & 20\% & 0\% & 0\% & 0\% & 2\% \\
Move Playingcard Away & 11\% & 22\% & 14\% & 0\% & 0\% & 3\% & 37\% \\
Move Stapler Pad & 0\% & 2\% & 4\% & 0\% & 0\% & 0\% & 0\% \\
Open Laptop & 32\% & 46\% & 3\% & 0\% & 0\% & 7\% & 12\% \\
Open Microwave & 20\% & 50\% & 14\% & 0\% & 0\% & 22\% & 23\% \\
Pick Diverse Bottles & 0\% & 6\% & 14\% & 0\% & 0\% & 1\% & 34\% \\
Pick Dual Bottles & 13\% & 12\% & 20\% & 0\% & 0\% & 1\% & 25\% \\
Place A2B Left & 1\% & 1\% & 12\% & 0\% & 0\% & 2\% & 18\% \\
Place A2B Right & 1\% & 6\% & 6\% & 0\% & 0\% & 0\% & 13\% \\
Place Bread Basket & 2\% & 4\% & 38\% & 0\% & 0\% & 1\% & 23\% \\
Place Bread Skillet & 1\% & 1\% & 19\% & 0\% & 0\% & 0\% & 34\% \\
Place Burger Fries & 27\% & 4\% & 45\% & 0\% & 0\% & 18\% & 37\% \\
Place Can Basket & 6\% & 5\% & 7\% & 0\% & 0\% & 2\% & 28\% \\
Place Cans Plasticbox & 5\% & 2\% & 27\% & 0\% & 0\% & 3\% & 23\% \\
Place Container Plate & 17\% & 45\% & 58\% & 1\% & 0\% & 1\% & 45\% \\
Place Dual Shoes & 4\% & 0\% & 3\% & 0\% & 0\% & 0\% & 19\% \\
Place Empty Cup & 7\% & 11\% & 53\% & 0\% & 0\% & 1\% & 8\% \\
Place Fan & 2\% & 10\% & 8\% & 0\% & 0\% & 1\% & 12\% \\
Place Mouse Pad & 0\% & 1\% & 11\% & 0\% & 0\% & 1\% & 20\% \\
Place Object Basket & 17\% & 2\% & 25\% & 0\% & 0\% & 0\% & 45\% \\
Place Object Scale & 0\% & 0\% & 20\% & 0\% & 0\% & 0\% & 35\% \\
Place Object Stand & 5\% & 11\% & 45\% & 0\% & 0\% & 0\% & 6\% \\
Place Phone Stand & 6\% & 7\% & 7\% & 0\% & 0\% & 2\% & 25\% \\
Place Shoe & 7\% & 6\% & 41\% & 0\% & 0\% & 2\% & 53\% \\
Press Stapler & 24\% & 29\% & 72\% & 6\% & 0\% & 3\% & 45\% \\
Put Bottles Dustbin & 4\% & 13\% & 10\% & 1\% & 0\% & 21\% & 35\% \\
Put Object Cabinet & 18\% & 18\% & 7\% & 0\% & 0\% & 1\% & 0\% \\
Rotate QRcode & 5\% & 15\% & 10\% & 0\% & 0\% & 1\% & 34\% \\
Scan Object & 1\% & 1\% & 7\% & 0\% & 0\% & 1\% & 23\% \\
Shake Bottle Horizontally & 51\% & 51\% & 96\% & 4\% & 18\% & 25\% & 44\% \\
Shake Bottle & 45\% & 60\% & 94\% & 10\% & 8\% & 19\% & 45\% \\
Stack Blocks Three & 0\% & 0\% & 14\% & 0\% & 0\% & 0\% & 18\% \\
Stack Blocks Two & 2\% & 1\% & 33\% & 0\% & 0\% & 0\% & 44\% \\
Stack Bowls Three & 17\% & 24\% & 24\% & 0\% & 0\% & 5\% & 87\% \\
Stack Bowls Two & 30\% & 41\% & 52\% & 0\% & 0\% & 6\% & 45\% \\
Stamp Seal & 0\% & 4\% & 15\% & 0\% & 0\% & 0\% & 0\% \\
Turn Switch & 15\% & 23\% & 20\% & 2\% & 1\% & 8\% & 16\% \\
\midrule
Average & 13.72\% & 16.34\% & 25.68\% & 1.74\% & 0.64\% & 4.96\% & 28.40\% \\
\end{longtable}
\endgroup
\FloatBarrier

\subsection{Public SimplerEnv Results}
\label{sec:full_simplerenv}

\begin{table}[!htbp]
\centering
\caption{Public SimplerEnv WidowX Visual Matching results used as the supplementary simulator comparison \citep{li24simpler}. This table is intentionally compact because SimplerEnv serves as public-reference context rather than the primary mechanism study. Baseline references include OpenVLA \citep{kim2024openvla}, CogACT \citep{li2024cogact}, SpatialVLA \citep{qu2025spatialvla}, $\pi_0$ and $\pi_0$-FAST \citep{black2024pi_0, pertsch2025fast}, GR00T-N1.5 \citep{bjorck2025gr00t}, Magma \citep{yang2025magma}, and StarVLA-$\alpha$ variants \citep{community2026starvla, ye2026starvla}. For NoTVLA, the success average is 65.7\% and the grasp average is 82.2\%.}
\label{tab:full_simplerenv}
\small
\begin{tabularx}{\linewidth}{l >{\raggedright\arraybackslash}X c}
\toprule
\textbf{Method} & \textbf{Backbone / Family} & \textbf{WidowX VM Avg.} \\
\midrule
OpenVLA & token VLA & 4.2\% \\
CogACT & hybrid policy & 51.3\% \\
SpatialVLA & spatial token VLA & 42.7\% \\
$\pi_0$ & VLA + flow expert & 27.1\% \\
$\pi_0$-FAST & tokenized $\pi$ variant & 48.3\% \\
GR00T-N1.5 & dual-system VLA & 61.9\% \\
Magma & multimodal foundation model & 35.8\% \\
Qwen-OFT & Qwen3-VL-4B + chunked continuous & 64.6\% \\
Qwen-PI & Qwen3-VL-4B + flow expert & 60.9\% \\
\textbf{NoTVLA} & Qwen3-VL-4B + narrative sparse action & 65.7\% \\
\bottomrule
\end{tabularx}
\end{table}
\FloatBarrier

\subsection{AGIBOT / Challenge-Style Results}
\label{sec:full_agibot}

\begin{table}[!htbp]
\centering
\caption{Official-server AGIBOT challenge-style results \citep{bu2025agibot}. These numbers are reported as an external benchmark context rather than as the primary evidence for the mechanism. UniVLA is included as the available public comparison \citep{bu2025univla}.}
\label{tab:agibot_full}
\resizebox{\linewidth}{!}{
\begin{tabular}{lccccccccccc}
\toprule
\textbf{Method} & \textbf{Total} & \textbf{1} & \textbf{2} & \textbf{3} & \textbf{4} & \textbf{5} & \textbf{6} & \textbf{7} & \textbf{8} & \textbf{9} & \textbf{10} \\
\midrule
UniVLA & 2.795 & 0.097 & 0.020 & 0.033 & 0.350 & 0.260 & 0.400 & 1.000 & 0.080 & 0.375 & 0.180 \\
\textbf{NoTVLA} & 3.697 & 0.161 & 0.304 & 0.350 & 0.320 & 0.448 & 0.510 & 0.600 & 0.262 & 0.350 & 0.392 \\
\bottomrule
\end{tabular}
}
\end{table}
\FloatBarrier

\subsection{Semantic Probe Details}
\label{sec:semantic_probe_appendix}

\begin{table}[!htbp]
\centering
\caption{Training-step probe accuracy on task-grounded QA.}
\label{tab:probe_over_steps}
\resizebox{\linewidth}{!}{
\begin{tabular}{lcccccccc}
\toprule
\textbf{Model} & \textbf{0} & \textbf{100} & \textbf{200} & \textbf{300} & \textbf{500} & \textbf{1000} & \textbf{2000} & \textbf{4000} \\
\midrule
Qwen-PI & 97\% & 68\% & 23\% & 0\% & 0\% & 0\% & 0\% & 0\% \\
\textbf{NoTVLA} & 97\% & 88\% & 79\% & 80\% & 76\% & 76\% & 76\% & 76\% \\
\bottomrule
\end{tabular}
}
\end{table}
\FloatBarrier

\begin{table}[!htbp]
\centering
\caption{Detailed semantic OOD task table with explicit Qwen-PI matched comparison. This compact table reports representative task-level entries; the aggregate OOD categories in Table~\ref{tab:semantic_ood} are averaged separately over inverse ordering, unseen color composition, choice-based placement, and novel concept transfer.}
\label{tab:semantic_ood_full}
\resizebox{\linewidth}{!}{
\begin{tabular}{lcccccccc}
\toprule
\textbf{Task} & \textbf{$\pi_0$-30k} & \textbf{$\pi_0$-5k} & \textbf{$\pi_{0.5}$-30k} & \textbf{$\pi_{0.5}$-5k} & \textbf{Qwen-PI-30k} & \textbf{Qwen-PI-5k} & \textbf{NoTVLA-30k} & \textbf{NoTVLA-5k} \\
\midrule
Place mouse pad & 23\% & 22\% & 35\% & 32\% & 27\% & 22\% & 25\% & 23\% \\
Stack blocks two & 90\% & 65\% & 93\% & 73\% & 96\% & 79\% & 90\% & 87\% \\
Stack blocks two (inverse) & 0.0\% & 0.0\% & 0.0\% & 0.0\% & 0.0\% & 0.0\% & 63\% & 70\% \\
Stack random color blocks & 30\% & 18\% & 35\% & 23\% & 48\% & 58\% & 62\% & 69\% \\
Place block flagpad & 32\% & 55\% & 44\% & 59\% & 57\% & 62\% & 84\% & 78\% \\
\bottomrule
\end{tabular}
}
\end{table}
\FloatBarrier

\subsection{Camera and Depth Robustness}
\label{sec:depth_noise}

\begin{table}[!htbp]
\centering
\caption{Robustness analysis against depth-anchor noise. Gaussian noise with the listed maximum amplitude is injected into the depth anchor, and each task is evaluated over 20 trials. The 0\,cm row uses this stress-test protocol and trial count, so it is not expected to exactly match the 50-trial ablation averages in Table~\ref{tab:camera_depth}.}
\label{tab:depth_noise_robustness}
\resizebox{\linewidth}{!}{
\begin{tabular}{lcccc}
\toprule
\multirow{2}{*}{\textbf{Task}} & \multicolumn{4}{c}{\textbf{Noise Level ($\sigma_{\max}$)}} \\
\cmidrule(lr){2-5}
& \textbf{0 cm} & \textbf{1 cm} & \textbf{10 cm} & \textbf{100 cm} \\
\midrule
Place a2b left & 40\% & 35\% & 20\% & 10\% \\
Place a2b right & 30\% & 25\% & 15\% & 5\% \\
Place object scale & 65\% & 25\% & 25\% & 15\% \\
Stack blocks three & 35\% & 30\% & 0\% & 0\% \\
Stack bowl three & 65\% & 40\% & 10\% & 10\% \\
Blocks ranking size & 20\% & 20\% & 0\% & 0\% \\
Blocks ranking rgb & 40\% & 35\% & 5\% & 10\% \\
\midrule
\textbf{Average} & \textbf{42\%} & \textbf{30\%} & \textbf{11\%} & \textbf{7\%} \\
\bottomrule
\end{tabular}
}
\end{table}

Small depth perturbations degrade performance but do not immediately break the policy, which indicates that closed-loop re-planning can correct moderate grounding error. Large perturbations cause a steep drop, especially on stacking and ranking tasks, confirming that accurate depth is a central part of the NoTVLA interface rather than an optional sensor detail.

\begin{table}[!htbp]
\centering
\caption{Camera robustness under mild fisheye distortion. Anchor-based grounding is compared with direct UVD prediction under the same distorted image condition.}
\label{tab:camera_shift_full}
\resizebox{\linewidth}{!}{
\begin{tabular}{lccc}
\toprule
\textbf{Task} & \textbf{Original Camera + Depth} & \textbf{Fisheye + Anchor Depth} & \textbf{Fisheye + Direct UVD} \\
\midrule
Place a2b left (clean) & 39\% & 32\% & 16\% \\
Place a2b right (clean) & 28\% & 24\% & 13\% \\
Place a2b randomly (hard) & 36\% & 30\% & 27\% \\
Place a2b randomly (clean) & 34\% & 25\% & 23\% \\
\midrule
\textbf{Average} & \textbf{34.3\%} & \textbf{27.8\%} & \textbf{19.8\%} \\
\bottomrule
\end{tabular}
}
\end{table}
\FloatBarrier

\subsection{Ablations}
\label{sec:ablations_appendix}

\begin{table}[!htbp]
\centering
\caption{Representation ablation summary over the seven tasks in Table~\ref{tab:camera_depth}. The same policy family is used while changing the sparse action grounding format.}
\label{tab:ablation_representation}
\small
\setlength{\tabcolsep}{5pt}
\resizebox{\linewidth}{!}{%
\begin{tabular}{lccc}
\toprule
\textbf{Variant} & \textbf{Object-Recognition Avg.} & \textbf{Long-Horizon Avg.} & \textbf{Overall Avg.} \\
\midrule
Sparse keyframes + direct UVD & 30.3\% & 52.3\% & 42.9\% \\
Sparse keyframes + anchor point only & 32.7\% & 46.3\% & 40.4\% \\
\textbf{Full NoTVLA} & \textbf{38.3\%} & \textbf{65.3\%} & \textbf{53.7\%} \\
\bottomrule
\end{tabular}%
}
\end{table}

In the anchor-point-only ablation, depth is removed from the narrative action interface and the detokenizer uses the current calibrated depth observation at each predicted image anchor to construct executable 3D targets. This isolates the value of explicitly predicting depth-related tokens while keeping the same projection pipeline.

The depth-sensitive grasping results are reported in Table~\ref{tab:depth_sensitive_grasp}; the appendix does not repeat the same table to avoid duplicate result reporting.
\FloatBarrier

\begin{table*}[!htbp]
\centering
\caption{Trajectory rendering quality for the deterministic detokenizer. The 10-point variant is slightly stronger, but the 5-point variant keeps the narrative action shorter while remaining close in quality. $\uparrow$ indicates higher is better and $\downarrow$ indicates lower is better. Baseline references include Magma \citep{yang2025magma}, HAMSTER \citep{li2025hamster}, and LLARVA \citep{niu2024llarva}; DTW and Hausdorff distance are standard trajectory distance diagnostics \citep{senin2008dynamic, rogers1998hausdorff}.}
\label{tab:detokenizer_quality}
\scriptsize
\setlength{\tabcolsep}{3.5pt}
\begin{tabularx}{0.92\textwidth}{>{\raggedright\arraybackslash}X *{5}{>{\centering\arraybackslash}X}}
\toprule
\textbf{Metric} & \textbf{NoTVLA 5-pt} & \textbf{NoTVLA 10-pt} & \textbf{Magma} & \textbf{HAMSTER} & \textbf{LLARVA} \\
\midrule
Cover F1 $\uparrow$ & 0.9546 & \textbf{0.9567} & 0.9282 & 0.9451 & 0.9189 \\
Cover precision $\uparrow$ & 0.9558 & \textbf{0.9612} & 0.9484 & 0.9504 & 0.9201 \\
Cover recall $\uparrow$ & 0.9550 & \textbf{0.9604} & 0.9534 & 0.9440 & 0.9415 \\
DTW $\downarrow$ & 0.4564 & \textbf{0.4554} & 0.8353 & 0.7945 & 0.4894 \\
Endpoint error $\downarrow$ & 0.0739 & \textbf{0.0729} & 0.1784 & 0.1181 & 0.0771 \\
Frechet $\downarrow$ & 0.1035 & \textbf{0.0945} & 0.2296 & 0.1311 & 0.1673 \\
Hausdorff $\downarrow$ & 0.0809 & \textbf{0.0795} & 0.1282 & 0.0822 & 0.0903 \\
Max orth. dist. $\downarrow$ & 0.0744 & \textbf{0.0725} & 0.1106 & 0.1232 & 0.0923 \\
Mean orth. dist. $\downarrow$ & 0.0322 & \textbf{0.0313} & 0.0476 & 0.0356 & 0.0677 \\
Median orth. dist. $\downarrow$ & 0.0286 & \textbf{0.0277} & 0.0439 & 0.0410 & 0.0532 \\
Startpoint error $\downarrow$ & 0.0778 & \textbf{0.0769} & 0.1449 & 0.1411 & 0.1743 \\
LCSS similarity $\uparrow$ & 0.9660 & \textbf{0.9671} & 0.9562 & 0.9496 & 0.9447 \\
\bottomrule
\end{tabularx}
\end{table*}
\FloatBarrier

\subsection{Real-Robot Details}
\label{sec:real_robot_details}

\begin{table}[!htbp]
\centering
\caption{Average execution time for successful real-robot trials. Failed trials are excluded from the time average.}
\label{tab:real_robot_success_time}
\resizebox{\linewidth}{!}{
\begin{tabular}{lccccc}
\toprule
\textbf{Method} & \textbf{Press Button} & \textbf{Place Cube Flag} & \textbf{Stack Cube Two} & \textbf{Insert Flower} & \textbf{Pick Fruit Bowl} \\
\midrule
DP3 & 13.6s & 46.2s & 60.4s & 65.6s & 40.6s \\
Qwen-PI & 15.9s & 60.6s & 70.6s & 75.6s & 50.6s \\
NoTVLA w/ direct UVD & 14.4s & 47.9s & 43.8s & 45.4s & 35.9s \\
\textbf{NoTVLA} & 26.3s & 62.5s & 65.5s & 61.5s & 45.5s \\
\bottomrule
\end{tabular}
}
\end{table}
\FloatBarrier

\subsection{Deferred Diagnostics}
\label{sec:deferred_diagnostics}

The current evaluation focuses on the evidence needed to test the central claim: a sparse narrative action interface can preserve more task-relevant semantics during robot adaptation and can improve semantic out-of-distribution manipulation under the tested protocols. Several diagnostics would further refine the mechanism analysis. In particular, future versions should add IK invalid or out-of-workspace rates, explicit no-merge closed-loop recovery ablations, and failure-mode counts separating wrong target grounding, depth error, interpolation failure, and late recovery. These diagnostics would help quantify where the interface fails; the present evidence chain distinguishes representation choice, semantic retention, grounding robustness, and deployment cost, but does not exhaustively explain every failure mode.

\section{Limitations and Scope}
\label{sec:limitations}

NoTVLA is designed around a semantics-preserving adaptation goal, not around maximizing performance on every manipulation regime. The method is most appropriate when a task benefits from retaining compositional language and visual semantics while delegating smooth execution to a transparent motion-rendering stage. It is less suited to regimes where success depends primarily on high-bandwidth contact dynamics, tactile feedback, very fast reactive control, or precise force regulation. The reported results should therefore be read as evidence for this interface choice in object-centric manipulation settings, not as a claim of universal VLA superiority.

The most important limitation is depth dependence. Anchor-conditioned grounding improves robustness compared with direct UVD prediction in our ablations, but NoTVLA still requires the depth value associated with the predicted anchor to be accurate enough for downstream projection and execution. The depth-noise results in Section~\ref{sec:depth_noise} show that small perturbations can be partially corrected by replanning, while large errors substantially reduce success. This also means that the method inherits calibration, occlusion, reflective-surface, and sensor-range failures from the depth source.

A second limitation is inference latency. In the current implementation, the VLM replans at a low narrative frequency and the deterministic detokenizer executes at a much higher actuator frequency. This separation makes the approach usable in the tested settings, but highly dynamic scenes may require faster multimodal inference, cached computation, or a learned recovery module. We therefore interpret the current latency as a systems bottleneck rather than a solved deployment problem.

Finally, the real-robot study is intentionally limited in scale. It is used to verify that the interface can be executed on hardware and that the latency/control-frequency separation is meaningful, not as comprehensive evidence of broad real-world generalization. The strongest evidence in this paper comes from matched simulation protocols, semantic retention probes, semantic OOD manipulation, and targeted grounding ablations; broader hardware coverage, multi-seed training, confidence intervals, and larger real-world task suites remain future work.

\section{Broader Impact and Responsible Use}
\label{sec:broader_impact}

NoTVLA is intended as a research framework for studying how robot adaptation can preserve semantic competence in VLM-based policies. Potential positive impacts include more reusable robot policies, more interpretable action interfaces, and reduced dependence on embodiment-specific low-level supervision. At the same time, any improvement in general-purpose robot manipulation can increase the risk of unsafe deployment if systems are used outside tested environments or without appropriate physical safeguards.

The current system is not designed for unsupervised deployment in safety-critical settings. It depends on calibrated sensors, bounded workspaces, and task-specific termination criteria. We recommend using the method with conventional robot safety mechanisms, workspace constraints, and human oversight when transferring beyond the reported benchmarks. The real-robot results in this paper are presented as limited deployment checks rather than evidence that the system is ready for broad autonomous use.

\section{Data and Implementation Details}
\label{app:data_implementation}

\subsection{Comparison Protocol and Claim Scope}
\label{app:comparison_scope}

We separate the experimental comparisons into three categories. First, \emph{matched-protocol comparisons} use the same simulator task definitions, evaluation splits, and success criteria as NoTVLA. These comparisons are the primary basis for claims about action-interface design. Second, \emph{backbone-family comparisons} compare NoTVLA with Qwen-style dense, chunked, or flow-based VLA variants. These results are used to study the effect of the adaptation interface under a closely related VLM family, while acknowledging that exact parameter count, pretraining version, and implementation details may differ across public systems. Third, \emph{public-reference comparisons} report published or public-server numbers to contextualize performance on community benchmarks such as RoboTwin and SimplerEnv.

The central claim of the paper does not require NoTVLA to outperform every public method on every benchmark. Instead, the claim is that changing the robot adaptation target from dense low-level control to a sparse narrative action interface can preserve more task-grounded semantic competence and that this preservation is associated with stronger semantic OOD manipulation. Backbone differences are a possible confound for aggregate benchmark tables, especially when Qwen2.5-VL-7B and Qwen3-VL-4B results are shown together. For this reason, we treat semantic retention probes, semantic OOD tasks, and anchor/depth ablations as the primary evidence for the mechanism, and use public benchmark and real-robot results as external validation and deployment context rather than as standalone causal proof.

\subsection{Data Sources and Coverage}
\label{app:data_sources}

We organize training and evaluation around three sources: RoboTwin-style simulated manipulation \citep{chen2025robotwin}, AGIBOT-style challenge data \citep{bu2025agibot}, and private real-robot demonstrations collected under matched task definitions. RoboTwin is used for the main controlled manipulation study because it provides a broad set of object-centric tasks with official clean/random evaluation settings, reported in our tables using the corresponding Easy/Hard naming convention. SimplerEnv is used as an additional public simulator reference. AGIBOT-style results are included as challenge-style evaluation from the official server rather than as the primary mechanism study. Real-robot tasks are used as a deployment sanity check.

RoboTwin training follows the official task setting with 50 clean demonstrations per task, using the official trajectory filtering and evaluation split. We report NoTVLA on 50 official RoboTwin tasks; several public baselines are available only for the 39-task public snapshot, which is marked with dashes in the task-level tables. The semantic OOD suite uses 9 custom tasks derived from held-out instruction compositions, including inverse ordering, unseen color composition, target choice, and novel concept transfer. Each semantic OOD task is evaluated with 100 trials. The real-robot study uses 20 demonstrations and evaluates each task-method pair with 20 trials.

\subsection{Statistical Reporting}
\label{app:statistical_reporting}

Most manipulation results are reported as success proportions over finite evaluation rollouts. RoboTwin, semantic OOD, anchor/depth ablation, and simulated card-grasping results use 100 trials per task-method setting. SimplerEnv follows the same trial-count convention as the corresponding public benchmark results. Real-robot task-method pairs use 20 trials. Aggregate success values are reported as unweighted averages over tasks; in particular, the semantic OOD average is the unweighted mean over the four OOD categories, and the real-robot summary averages over the five hardware tasks.

For proportion-valued success rates, uncertainty can be summarized with binomial confidence intervals, such as Wilson 95\% intervals. We do not include these intervals in the tables to keep the reporting compact, and we do not include multi-seed training variance. The reported numbers should therefore be interpreted as single-checkpoint, multi-rollout estimates rather than full training-and-evaluation uncertainty. This distinction is especially important for the real-robot study, where 20-trial estimates have wide intervals and are used as deployment sanity checks rather than primary evidence for broad real-world generalization.

Accordingly, we interpret large and protocol-matched gaps, together with semantic retention and ablation evidence, as the main support for the mechanism claim. The key semantic OOD comparison is NoTVLA against Qwen-PI, $\pi_0$, and $\pi_{0.5}$ style baselines under the same task suite. Small differences in public-reference benchmark tables are not treated as statistically significant head-to-head comparisons.

\subsection{Semantic Probe Construction}
\label{app:semantic_probe_details}

The task-grounded QA probe is designed to test whether robot fine-tuning preserves the VLM's ability to parse task-relevant visual and linguistic semantics. Each probe instance contains the current observation, the task instruction, and a short question about the target object, attribute, relation, or intended manipulation. Example question types include identifying the target object, selecting the correct color or attribute, determining the requested ordering relation, and resolving choice-based placement instructions.

Probe questions are generated from the same task family used for robot adaptation but are evaluated as language or short-answer predictions rather than as control rollouts. This makes the probe a diagnostic for semantic retention, not a substitute for manipulation performance. We therefore pair the probe with semantic OOD manipulation in the main text. The probe set contains 960 examples, balanced across five task categories. The question types cover target-object identification, color and attribute recognition, ordering relations, choice-based placement, and task-stage recognition. For example, a probe may ask ``What is your target object, bread or clock?'' with the answer ``bread''. Answers are scored by a rule-based regular-expression parser and reported as accuracy; all compared checkpoints are evaluated with the same prompt template and decoding settings.

\subsection{Action Representation and Keyframe Extraction}
\label{app:action_representation_details}

Dense demonstrations are converted into sparse narrative action sequences before fine-tuning. We select decision-relevant keyframes using gripper-state transitions and kinematic changes, then optionally insert sub-keyframes to preserve local geometric context. The default NoTVLA representation uses a fixed 5-point sparse waypoint format per narrative prediction. Variable-length extracted sequences are resampled to this format by retaining the endpoints and selecting intermediate waypoints at uniform arc-length intervals. Each waypoint contains image coordinates, depth-related tokens, gripper state, and rotation tokens.

The keyframe acceleration threshold in Eq.~\ref{eq:keyframe_cond} is set to 1.0\,m/s$^2$ for the main RoboTwin experiments and is applied to the translational component of the end-effector pose. The number of inserted sub-keyframes ranges from 2 to 10 before the final sparse formatting. Coordinate and depth tokens use a 0--1000 normalized token range for Qwen3-VL-4B experiments, while Qwen2.5-VL experiments use absolute image-coordinate values with the corresponding depth representation. Rotation is represented with Euler angles at the token level and converted to unit quaternions before SLERP-based interpolation. The gripper state is represented with open/close tokens. These choices are kept fixed for the representation ablations unless otherwise stated.

\subsection{Training Recipe}
\label{app:training_details}

NoTVLA fine-tunes an autoregressive VLM by next-token prediction over the narrative action sequence. We use Qwen3-VL-4B and Qwen2.5-VL-7B backbones across the reported experiments \citep{bai2025qwen2}. SimplerEnv and the semantic QA probe use Qwen3-VL-4B; RoboTwin and AGIBOT use Qwen2.5-VL-7B. Unless otherwise stated, NoTVLA is trained for 3 epochs using AdamW with learning rate $1\times10^{-5}$, batch size 32, and weight decay $1\times10^{-6}$. We do not use a separate learning-rate scheduler unless otherwise noted. The main training runs use 4 A100 GPUs, with peak training memory around 96\% of device memory.

For Qwen-PI and Qwen-OFT style baselines, we follow the corresponding StarVLA public setting \citep{community2026starvla, ye2026starvla}. Qwen-PI uses a Qwen3-VL-4B backbone with a flow-style action interface, while Qwen-OFT uses a Qwen3-VL-4B backbone with chunked continuous actions. The -30k and -5k suffixes denote the training-step checkpoints used for semantic OOD and retention probes. RoboTwin comparisons use the same 50-clean-demonstration setting where available. SimplerEnv comparisons for other methods are public-reference results rather than fully reproduced matched runs. Public-reference methods are cited as benchmark context and are not used as the sole basis for mechanism claims.

At inference time, each NoTVLA narrative decision contains roughly 50 generated tokens for the 5-point waypoint format. Narrative plans are refreshed at approximately 0.25--0.5\,Hz, corresponding to a 2--4\,s high-level replanning period in the current implementation, while the deterministic detokenizer produces commands at 80\,Hz. The main deployment experiments use peak inference memory of approximately 13\,GB.

\subsection{Deployment Setup}
\label{app:deployment}

The real-robot setup consists of a Franka arm with a Franka Panda gripper. We use an Intel RealSense D435if camera for RGB and depth input \citep{keselman2017intel}. Camera intrinsics and extrinsics are estimated using QR-code-based calibration. The predicted anchor point is paired with the corresponding depth value and projected into the robot base frame using the calibrated camera model.

Sparse narrative waypoints are de-quantized into 3D end-effector targets. Translation is rendered with cubic B-spline interpolation and orientation is rendered with SLERP. The controller executes the resulting trajectory at 80\,Hz, while high-level VLM replanning runs asynchronously at approximately 0.25--0.5\,Hz. When a new plan arrives during execution, the trajectory-merging procedure described in Section~\ref{sec:detokenization} aligns the pending waypoints with the current robot state and connects them through a smooth transition segment.

Real-robot trials use task-specific termination criteria. The insertion task is counted as a failure if the flower falls onto the table; other tasks are counted as failures if the robot does not complete the specified goal before the evaluation cutoff or if execution violates the task-specific safety condition. Successful-trial execution time excludes failed trials, matching the reporting convention in Table~\ref{tab:real_robot_success_time}.

\section{More Results and Reality Performance}

\begin{figure}
    \centering
    \includegraphics[width=1\linewidth]{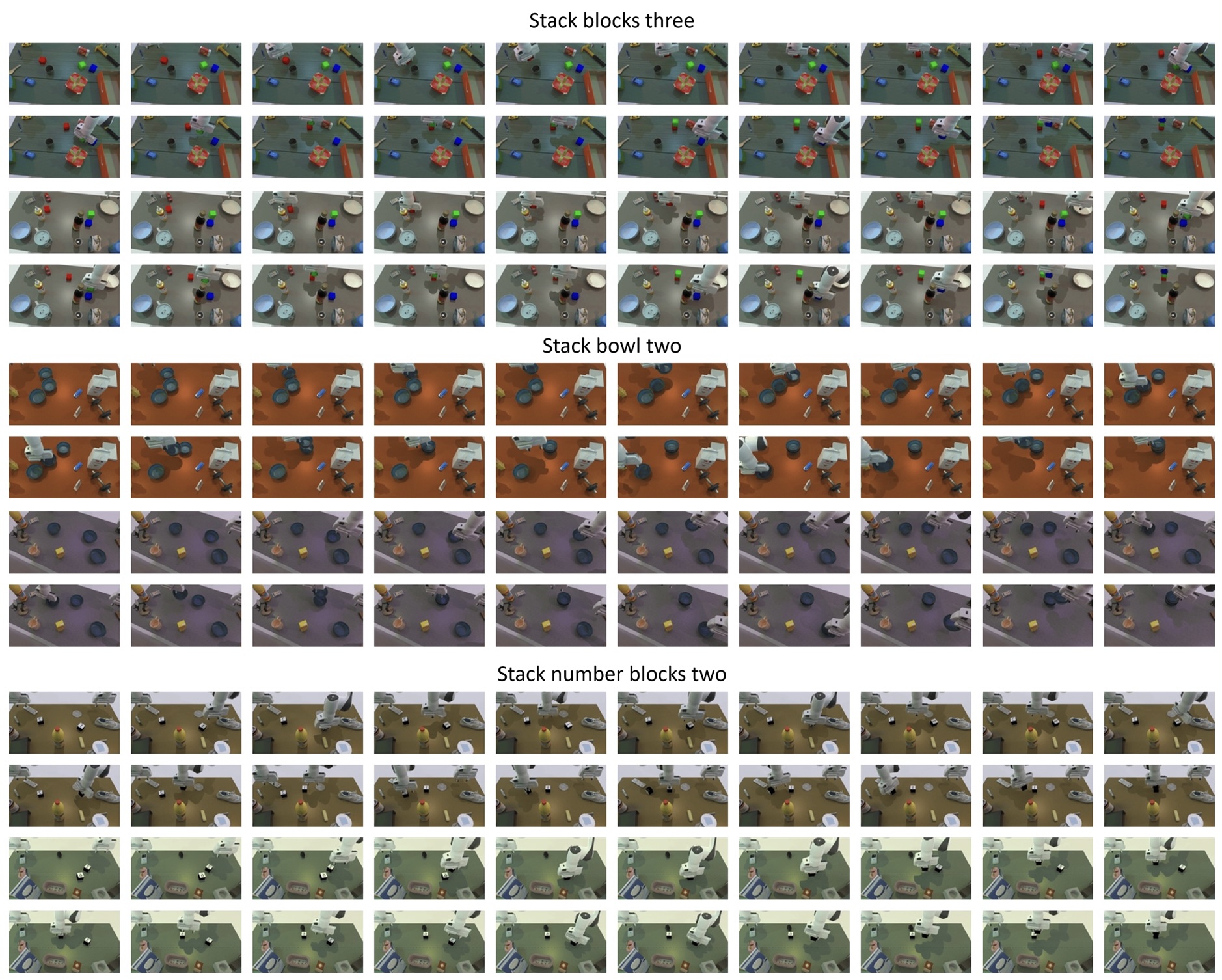}
    \caption{NoTVLA operation in simulation.}
    \label{fig:sim}
\end{figure}

\begin{figure}
    \centering
    \includegraphics[width=1\linewidth]{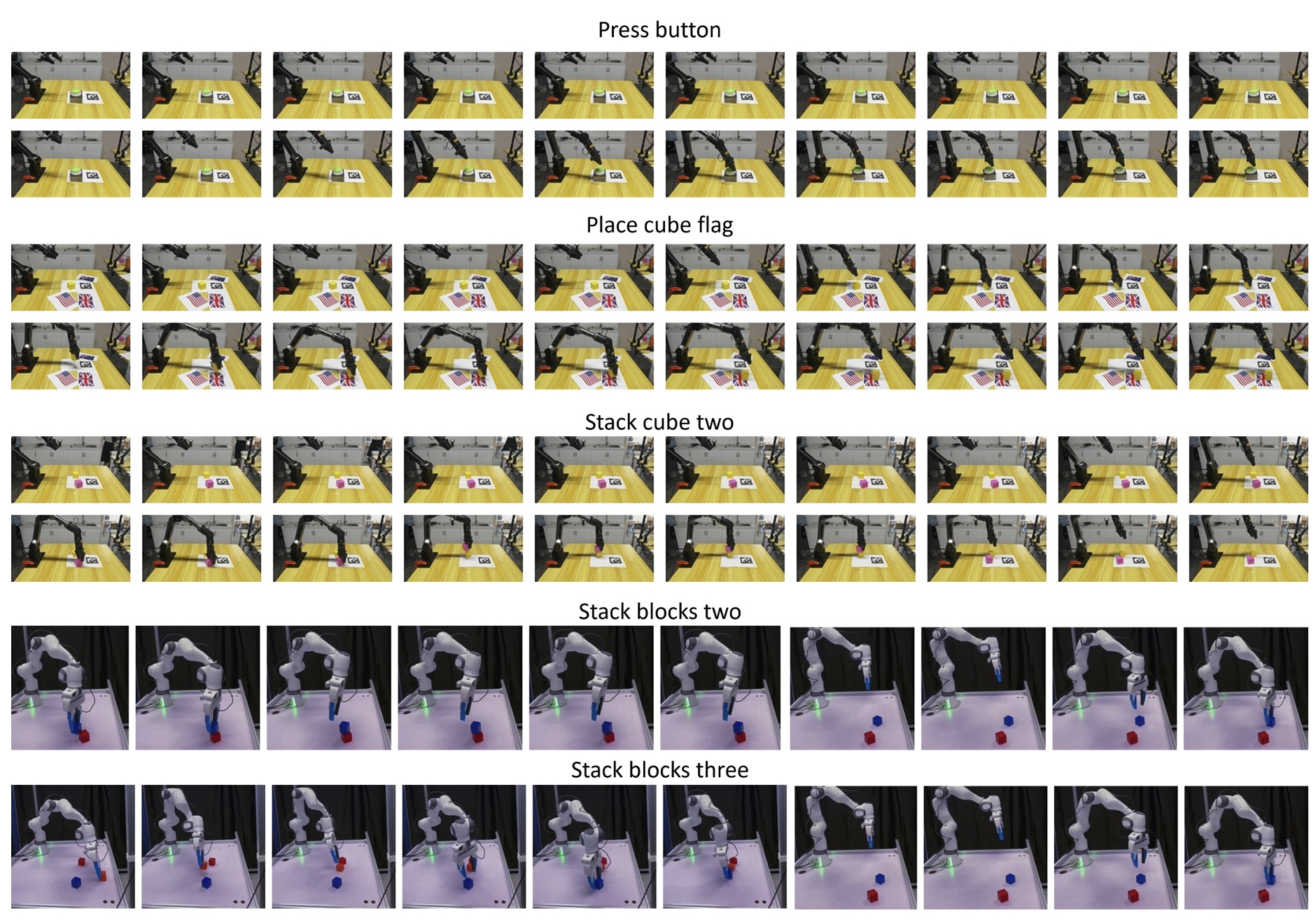}
    \caption{NoTVLA operation in reality.}
    \label{fig:real}
\end{figure}

NoTVLA maintains strong performance even when the main camera pose is randomized, as illustrated in Figure~\ref{fig:sim}. The same policy also transfers effectively to the real-robot setting, with representative executions shown in Figure~\ref{fig:real}.

\section{Task Descriptions}
\subsection{AGIBOT Challenge}
\label{sec:agibot_tasks}

\begin{description}
  \item[Clear the countertop waste - task 1] Remove loose trash and food scraps from the countertop, place them into the appropriate waste or recycling container, and wipe the surface clean.

  \item[Open drawer and store items - task 2] Open the specified drawer, place the target items inside following organizational guidelines, arrange them neatly, and close the drawer securely.

  \item[Heat the food in the microwave - task 3] Place the food in a microwave-safe container, set the appropriate power and time, start the microwave, and carefully remove and check the food temperature when finished.

  \item[Pack moving objects from conveyor - task 4] Identify and pick moving objects from the conveyor belt, place them into designated packing boxes or trays while maintaining speed and stability.

  \item[Pickup items from the freezer - task 5] Open the freezer, locate and retrieve specified frozen items using proper protection (e.g., gloves), move them to the target location, and close the freezer.

  \item[Restock supermarket items - task 6] Retrieve products from stock, refill shelves according to merchandising rules (facing front, orderly arrangement), replace expired or mislabeled items as needed.

  \item[Pack in the supermarket - task 7] Pack purchased goods for a customer by grouping items sensibly, protecting fragile goods, and arranging bags or boxes for safe and easy carrying.

  \item[Make a sandwich - task 8] Prepare bread and fillings according to the recipe, assemble ingredients in order, cut if required, and present a clean, ready-to-eat sandwich.

  \item[Clear table in the restaurant - task 9] Remove dishes, utensils, and leftovers from the table, clear and sanitize the surface, and reset place settings if necessary for the next guest.

  \item[Stamp the seal - task 10] Retrieve the stamp, align it with the required location on the document, press firmly to produce a clear impression, and verify alignment and clarity.
\end{description}

\subsection{RoboTwin Official Tasks}
\label{sec:official_tasks}

\begin{description}
  \item[Beat Block Hammer] Use a hammer to strike the block target repeatedly until the action is completed or the block is secured.

  \item[Blocks Ranking RGB] Rank blocks by visual RGB features and sort them according to the specified criterion.

  \item[Blocks Ranking Size] Rank blocks by size and arrange them in the required order.

  \item[Click Alarmclock] Locate the alarm clock and press or click the designated button to silence or activate the alarm.

  \item[Click Bell] Press the bell actuator at the indicated position to produce a ringing signal.

  \item[Dump Bin Bigbin] Open the large bin and empty its contents into the designated disposal area.

  \item[Grab Roller] Reach for and grasp the roller object securely, then lift or move it to the target location.

  \item[Handover Mic] Grasp the microphone and transfer it to another agent or specified handover position.

  \item[Hanging Mug] Pick up a mug from a hanging position and place it on the table or target surface.

  \item[Lift Pot] Grasp the pot handle(s) and lift it carefully to the required height or location.

  \item[Move Can Pot] Move the can and the pot to their respective target positions as specified.

  \item[Move Pillbottle Pad] Pick up the pill bottle and place it on the target pad or designated area.

  \item[Move Playingcard Away] Remove the playing card from its current location and place it at the specified away location.

  \item[Move Stapler Pad] Pick up the stapler and position it on the given pad or surface.

  \item[Open Laptop] Open the laptop lid to a functional angle and ensure it is ready for use.

  \item[Open Micro Wave] Open the microwave door and prepare the interior for loading or unloading items.

  \item[Place A2B Left] Pick up object A and place it at location B on the left-side target as specified.

  \item[Place A2B Right] Pick up object A and place it at location B on the right-side target as specified.

  \item[Place Bread Basket] Place the bread item into the basket, arranging it neatly.

  \item[Place Bread Skillet] Place the bread onto the skillet or pan and position it correctly for cooking or serving.

  \item[Place Burger Fries] Pack or arrange the burger and fries together in the specified container.

  \item[Place Can Basket] Place the can into the basket at the designated spot.

  \item[Place Cans Plasticbox] Place the cans into the plastic box following placement constraints.

  \item[Place Container Plate] Put the container onto the plate or place the plate into the container as required.

  \item[Place Dual Shoes] Pick up the pair of shoes and place them together at the target location.

  \item[Place Empty Cup] Place an empty cup at the specified position without spilling or tilting.

  \item[Place Fan] Position the fan at the target location and orientation.

  \item[Place Mouse Pad] Place the mouse pad flat at the designated workspace area.

  \item[Place Object Basket] Place the specified object into the basket, ensuring stable placement.

  \item[Place Object Scale] Place the object on the scale platform for weighing.

  \item[Place Object Stand] Position the object on the stand securely and centered.

  \item[Place Phone Stand] Place the phone onto the stand in the correct orientation.

  \item[Place Shoe] Pick up the shoe and place it at the designated spot.

  \item[Press Stapler] Press down the stapler mechanism to staple the target documents or materials.

  \item[Put Object Cabinet] Place the object into the cabinet compartment and close the door if required.

  \item[Rotate QR Code] Rotate the QR code marker to the specified orientation for scanning.

  \item[Scan Object] Position the scanner or object and perform a scan to capture required data.

  \item[Shake Bottle] Grasp the bottle and shake it vertically or as specified to mix contents.

  \item[Shake Bottle Horiz.] Grasp the bottle and shake it horizontally to achieve the required motion.

  \item[Stack Block Three] Stack three blocks vertically in the specified order and ensure stability.

  \item[Stack Block Two] Stack two blocks as required and confirm alignment.

  \item[Stack Bowl Three] Stack three bowls together with stable nesting.

  \item[Stack Bowl Two] Stack two bowls together with proper alignment.

  \item[Stamp Seal] Align the seal with the document and press to produce a clear stamp impression.

  \item[Turn Switch] Rotate or flip the switch to the target position to change device state.
\end{description}

\subsection{RoboTwin Custom Tasks}
\label{sec:custom_tasks}

\begin{description}
  
  \item[Stack Blocks Two Inverse] Stack two blocks in the inverse of a previously seen order, requiring the policy to reinterpret the ordering instruction rather than replay the training distribution.

  \item[Stack Random Color Blocks] Following an unseen color-order instruction, stack two random color blocks in the required order.

  \item[Move Block Random Color Pad] Pick the colorful block and place it onto the correspondingly colored pad, matching color pairing and ensuring stable placement.
  \item[Stack Numberblocks Two] Pick the two specified number blocks and stack them in required numerical order, bottom to top.
  \item[Stack Random Blocks Two From Three] Given three candidate blocks, select the two specified by the instruction and stack them in the required color order.
  \item[Place Block Flagpad] Pick the target block and place it onto the flag-marked pad, centered and fully supported.
  \item[Move Colorful Block Colorful Pad] Pick the colorful block and place it onto the pad with the matching color pattern, aligning edges for stable contact.
  \item[Place Block Flagpad (choice)] From multiple pads, choose a flag-labeled pad and place the block there.
  \item[Place A2B Randomly] Pick an object from region A and place it at a random valid pose within region B, staying within bounds and avoiding obstacles.
\end{description}


\newpage

\end{document}